\documentclass[10pt]{article}
\usepackage{fullpage}
\usepackage{times,amsmath,amsfonts,amssymb,url,color,graphicx}

\usepackage[numbers]{natbib}

\usepackage{algorithm}
\usepackage{algorithmic}
\algsetup{indent=2em}

\usepackage{etaremune}

\usepackage{tikz}
\usetikzlibrary{arrows,decorations.pathmorphing,backgrounds,fit}
\usetikzlibrary{positioning} 

\newcommand{\sign}{\mathrm{sign}}

\newcommand{\reals}{\mathbb{R}}

\newcommand{\DNN}{\mathrm{DNN}}

\newcommand{\BlackBox}{\rule{1.5ex}{1.5ex}}  

\newcommand{\secref}[1]{Section~\ref{#1}}

\DeclareMathOperator*{\E}{\mathbb{E}}

\begin{document}

\title{Long-term Planning by Short-term Prediction}
\date{Mobileye}

\author{Shai Shalev-Shwartz \and Nir Ben-Zrihem \and Aviad Cohen \and Amnon Shashua}

\maketitle

\begin{abstract}
We consider planning problems, that often arise in autonomous driving applications, in which an agent should decide on immediate actions so as to optimize a long term objective. For example, when a car tries to merge in a roundabout it should decide on an immediate acceleration/braking command, while the long term effect of the command is the success/failure of the merge. Such problems are characterized by continuous state and action spaces, and by interaction with multiple agents, whose behavior can be adversarial. We argue that dual versions of the MDP framework (that depend on the value function and the $Q$ function) are problematic for autonomous driving applications due to the non Markovian of the natural state space representation, and due to the continuous state and action spaces.  We propose to tackle the planning task by decomposing the problem into two phases: First, we apply supervised learning for predicting the near future based on the present. We require that the predictor will be differentiable with respect to the representation of the present. Second, we model a full trajectory of the agent using a recurrent neural network, where unexplained factors are modeled as (additive) input nodes. This allows us to solve the long-term planning problem using supervised learning techniques and direct optimization over the recurrent neural network. Our approach enables us to learn robust policies by incorporating adversarial elements to the environment. 
\end{abstract}

\section{Introduction}
Two of the most crucial elements of autonomous driving systems are sensing and planning. Sensing deals with finding a compact representation of the present state of the environment, while planning deals with deciding on what actions to take so as to optimize future objectives. Supervised machine learning techniques are very useful for solving sensing problems. In this paper we describe a machine learning algorithmic framework for the planning part. Traditionally, machine learning approaches for planning are studied under the framework of Reinforcement Learning (RL) --- see \cite{bertsekas1995dynamic,kaelbling1996reinforcement,sutton1998reinforcement,szepesvari2010algorithms} for a general overview and \cite{kober2013reinforcement} for a comprehensive review of reinforcement learning in robotics. 

Typically, RL is performed in a sequence of consecutive rounds. At round $t$, the planner (a.k.a. the agent) observes a state, $s_t \in S$, which represents the agent as well as the environment. It then should decide on an action $a_t \in A$.  After performing the action, the agent receives an immediate reward, $r_t \in \reals$, and is moved to a new state, $s_{t+1}$. As a simple example, consider an adaptive cruise control (ACC) system, in which a self driving vehicle should implement acceleration/braking so as to keep an adequate distance to a preceding vehicle while maintaining smooth driving. We can model the state as a pair, $s_t = (x_t,v_t) \in \reals^2$, where $x_t$ is the distance to the preceding vehicle and $v_t$ is the velocity of the car relative to the velocity of the preceding vehicle. The action $a_t \in \reals$ will be the acceleration command (where the car slows down if $a_t < 0$). The reward can be some function that depends on $|a_t|$ (reflecting the smoothness of driving) and on $s_t$ (reflecting that we keep a safe distance from the preceding vehicle). The goal of the planner is to maximize the cumulative reward (maybe up to a time horizon or a discounted sum of future rewards). To do so, the planner relies on a policy, $\pi : S \to A$, which maps a state into an action.

Supervised Learning (SL) can be viewed as a special case of RL, in which  $s_t$ is sampled i.i.d. from some distribution over $S$ and the reward function has the form $r_t=-\ell(a_t,y_t)$, where $\ell$ is some loss function, and the learner observes the value of $y_t$ which is the (possibly noisy) value of the optimal action to take when viewing the state $s_t$. 

There are several key differences between the fully general RL model and the specific case of SL. These differences makes the general RL problem much harder. 
\begin{enumerate}
\item In SL, the actions (or predictions) taken by the learner have no effect on the environment. In particular, $s_{t+1}$ and $a_t$ are independent. This has two important implications:
\begin{itemize}
\item In SL we can collect a sample $(s_1,y_1),\ldots,(s_m,y_m)$ in advance, and only then search for a policy (or predictor) that will have good accuracy on the sample. In contrast, in RL, the state $s_{t+1}$ usually depends on the action (and also on the previous state), which in turn depends on the policy used to generate the action. This ties the data generation process to the policy learning process. 
\item Because actions do not effect the environment in SL, the contribution of the choice of $a_t$ to the performance of $\pi$ is local, namely, $a_t$ only affects the value of the immediate reward. In contrast, in RL, actions that are taken at round $t$ might have a long-term effect on the reward values in future rounds.
\end{itemize}
\item In SL, the knowledge of the ``correct'' answer, $y_t$, together with the shape of the reward, $r_t = -\ell(a_t,y_t)$, gives us a full knowledge of the reward for all possible choices of $a_t$. Furthermore, this often enables us to calculate the derivative of the reward with respect to $a_t$. In contrast, in RL, we only observe a ``one-shot'' value of the reward for the specific choice of action we took. This is often called a ``bandit'' feedback. It is one of the main reasons for the need of ``exploration'', because if we only get to see a ``bandit'' feedback, we do not always know if the action we took is the best one. 
\end{enumerate}

Before explaining our approach for tackling these difficulties, we briefly describe the key idea behind most common reinforcement learning algorithms. Most of the algorithms rely in some way or another on the mathematically elegant model of a Markov Decision Process (MDP), pioneered by the work of Bellman \cite{bellman1956dynamic,bellman1971introduction}. The Markovian assumption is that the distribution of $s_{t+1}$ is fully determined given $s_t$ and $a_t$. This yields a closed form expression for the cumulative reward of a given policy in terms of the stationary distribution over states of the MDP. The stationary distribution of a policy can be expressed as a solution to a linear programming problem. This yields two families of algorithms: optimizing with respect to the primal problem, which is called policy search, and optimizing with respect to the dual problem, whose variables are called the \emph{value function}, $V^\pi$. The value function determines the expected cumulative reward if we start the MDP from the initial state $s$, and from there on pick actions according to $\pi$. A related quantity is the state-action value function, $Q^\pi(s,a)$, which determines the cumulative reward if we start from state $s$, immediately pick action $a$, and from there on pick actions according to $\pi$. The $Q$ function gives rise to a crisp characterization of the optimal policy (using the so called Bellman's equation), and in particular it shows that the optimal policy is a deterministic function from $S$ to $A$ (in fact, it is the greedy policy with respect to the optimal $Q$ function). 

In a sense, the key advantage of the MDP model is that it allows us to couple all the future into the present using the $Q$ function. That is, given that we are now in state $s$, the value of $Q^\pi(s,a)$ tells us the effect of performing action $a$ at the moment on the entire future. Therefore, the $Q$ function gives us a local measure of the quality of an action $a$, thus making the RL problem more similar to SL. 

Most reinforcement learning algorithms approximate the $V$ function or the $Q$ function in one way or another. Value iteration algorithms, e.g. the $Q$ learning algorithm \cite{watkins1992q}, relies on the fact that the $V$ and $Q$ functions of the optimal policy are fixed points of some operators derived from Bellman's equation. Actor-critic policy iteration algorithms aim to learn a policy in an iterative way, where at iteration $t$, the ``critic'' estimates $Q^{\pi_t}$ and based on this, the ``actor'' improves the policy. 

Despite the mathematical elegancy of MDPs and the conveniency of switching to the $Q$ function representation, there are several limitations of this approach. First, as noted in \cite{kober2013reinforcement}, usually in robotics, we may only be able to find some approximate notion of a Markovian behaving state. Furthermore, the transition of states depends not only on the agent's action, but also on actions of other players in the environment. For example, in the ACC example mentioned previously, while the dynamic of the autonomous vehicle is clearly Markovian, the next state depends on the behavior of the other driver, which is not necessarily Markovian. One possible solution to this problem is to use partially observed MDPs~\cite{white1991survey}, in which we still assume that there is a Markovian state, but we only get to see an observation that is distributed according to the hidden state. A more direct approach considers game theoretical generalizations of MDPs, for example the Stochastic Games framework. Indeed, some of the algorithms for MDPs were generalized to multi-agents games. For example, the minimax-Q learning \cite{littman1994markov} or the Nash-Q learning \cite{hu2003nash}. Other approaches to Stochastic Games are explicit modeling of the other players, that goes back to Brown's fictitious play~\cite{brown1951iterative}, and vanishing regret learning algorithms \cite{hart2000simple,CesaBianchiLu06}.  See also \cite{uther1997adversarial, Thrun95a,kearns2002near,brafman2003r}. As noted in \cite{shoham2007if}, learning in multi-agent setting is inherently more complex than in the single agent setting.

A second limitation of the $Q$ function representation arises when we depart from a tabular setting. The tabular setting is when the number of states and actions is small, and therefore we can express $Q$ as a table with $|S|$ rows and $|A|$ columns. However, if the natural representation of $S$ and $A$ is as Euclidean spaces, and we try to discretize the state and action spaces, we obtain that the number of states/actions is exponential in the dimension. In such cases, it is not practical to employ the tabular setting. Instead, the $Q$ function is approximated by some function from a parametric hypothesis class (e.g. neural networks of a certain architecture). For example, the deep-Q-networks (DQN) learning algorithm of \cite{mnih2015human} has been successful at playing Atari games. In DQN, the state space can be continuous but the action space is still a small discrete set. There are approaches for dealing with continuous action spaces (e.g. \cite{silver2014deterministic}), but they again rely on approximating the $Q$ function. In any case, the $Q$ function is usually very complicated and sensitive to noise, and it is therefore quite hard to learn it. Indeed, it was observed that value based methods rarely work out-of-the-box in robotic applications~\cite{kober2013reinforcement}, and that the best performing methods rely on a lot of prior knowledge and reward shaping \cite{laud2004theory,ng1999policy}. Intuitively, the difficulty in learning $Q$ is that we need to implicitly understand the dynamics of the underlying Markov process. 

In the autonomous driving domain we tackle in this paper, the multi-agent adversarial environment leads to non-Markovianity of the natural state representation. Moreover, the natural state and action representations are continuous in nature. Taken together, we found out that $Q$-based learning approaches rarely work out-of-the-box, and require long training time and advanced reward shaping.  

A radically different approach has been introduced by Schmidhuber~\cite{schmidhuber1991reinforcement}, who tackled the RL problem using a recurrent neural network (RNN). Following \cite{schmidhuber1991reinforcement}, there have been several additional algorithms that rely on RNNs for RL problems. For example,  Backer~\cite{bakker2001reinforcement} proposed to tackle the RL problem using recurrent networks with the LSTM architecture. His approach still relies on the value function. Sch{\"a}fer~\cite{schafer2008reinforcement} used RNN to model the dynamics of partially observed MDPs. Again, he still relies on explicitly modeling the Markovian structure. There have been few other approaches to tackle the RL problem without relying on value functions. Most notably is the REINFORCE framework of Williams~\cite{williams1992simple}. It has been recently successful for visual attention \cite{mnih2014recurrent,xu2015show}. As already noted by \cite{schmidhuber1991reinforcement}, the ability of REINFORCE to estimate the derivative of stochastic units can be straightforwardly combined within the RNN framework. 

In this paper we combine Schmidhuber's approach, of tackling the policy learning problem directly using a RNN, with the notions of multi-agents games and robustness to adversarial environments from the game theory literature. Furthermore, we do not explicitly rely on any Markovian assumption. Our approach is described in the next section.

\section{Planning by Prediction}

Throughout, we assume that the state space, $S$, is some subset of $\reals^d$, and the action space, $A$, is some subset of $\reals^k$. This is the most natural representation in many applications, and in particular, the ones we describe in \secref{sec:experiments}. 

Our goal is to learn a policy $\pi: S \rightarrow A$. As is standard in machine learning, we bias ourselves to pick a policy function $\pi$ from a hypothesis class $\mathcal{H}$. Namely, $\mathcal{H}$ is a predefined set of policy functions from which we should pick the best performing one. 
In order to learn $\pi$ using the SL framework, one would need a training set of pairs (state,optimal-action). We of course do not have such a training set. Instead, we only have an access to a ``simulator'', that can be used to assess the quality of $\pi$. Formally, fixing a horizon $T$, any policy $\pi$ induces a distribution over $\reals^T$, such that the probability of  $(r_1,\ldots,r_T) \in \reals^T$ is the probability to apply our simulator for $T$ steps, while on step $t$ we observe $s_t$, feed the simulator with the action $a_t=\pi(s_t)$, and observe the reward $r_t$. Denote by $B$ the random bits used by the simulator, we note that we can write the vector $r = (r_1,\ldots,r_T)$ of rewards as a deterministic function $R(B,\pi)$. We use $R_t(B,\pi)$ to denote the $t$'th element of $R(B,\pi)$. We can now formulate the problem of learning the policy $\pi$ as the following optimization problem:
\begin{equation} \label{eqn:policyOptimization}
\max_{\pi \in \mathcal{H}} ~ \E_B\left[  \sum_{t=1}^T R_t(B,\pi) \right] ~,
\end{equation}
where the expectation is over the distribution over $B$.

We assume that the hypothesis class, $\mathcal{H}$, is the set of deep neural networks (DNN) of a certain predefined architecture, and therefore every $\pi \in \mathcal{H}$ is parametrized by a vector of weights, $\theta$. We use $\pi_\theta$ to denote the policy associated with the vector $\theta$.  

If we could express $R(B,\pi_\theta)$ as a differential function of $\theta$, we could have utilized the Stochastic Gradient Descent (SGD) approach for maximizing \eqref{eqn:policyOptimization}. That is, starting with an initial $\theta$, at each iteration of SGD we first sample $B$, then we calculate the gradient of $\sum_{t=1}^T R_t(B,\pi_\theta)$ with respect to $\theta$, and finally we update $\theta$ based on this gradient.

Our key observation is that by solving two SL problems, described below, we can approximate $R(B,\pi_\theta)$ by a differential function of $\theta$. Hence, we can implement SGD for learning $\pi_\theta$ directly.

The goal of the first SL problem is to learn a deep neural network (DNN), that represents the mapping from a (state,action) pair into the immediate reward value. We denote this DNN by $\DNN_r$ and it is formally described as a function $\DNN_r : S \times A \to \reals$. We shall later explain how to learn $\DNN_r$ using SL, but for now lets just assume that we can do it and have the network $\DNN_r$ such that $\DNN_r(s_t,a_t) \approx r_t$. 
The goal of the second SL problem is to learn a DNN that represents the mapping from (state,action) into the next state. Formally, this DNN is the function $\DNN_N : S \times A \rightarrow S$, and for now lets assume we managed to learn $\DNN_N$ in a supervised manner such that $\DNN_N(s_t,a_t) \approx s_{t+1}$. 

Equipped with $\DNN_r$ and $\DNN_N$ we can describe the process of generating a random $B$ and calculating $R(B,\pi_\theta)$ as follows. Initially, the simulator picks a seed for its pseudo random number generator and then it determines the initial state $s_1$. At round $t$, the agent receives $s_t$ from the simulator and applies $\pi_\theta$ to generate the action $a_t = \pi_\theta(s_t)$. The simulator receives $a_t$ and generates $r_t$ and $s_{t+1}$. At the same time, the agent applies $\DNN_r$ to generate $\hat{r}_t = \DNN_r(s_t)$ and applies $\DNN_N$ to generate $\hat{s}_{t+1} = \DNN_N(s_t)$. Let us denote $\nu_{t+1}= s_{t+1} - \hat{s}_{t+1}$. Therefore, if the simulator receives $\hat{s}_{t+1}$ it can generate $\nu_{t+1}$ and send it to the agent. 

A single round of this process is depicted below. The entire process is obtained by repeating the shaded part of the picture $T$ times. Solid arrows represent differentiable propagation of information while dashed arrows represent non-differentiable propagation of information. 

\begin{center}
\begin{tikzpicture}[var/.style={minimum size=12mm,circle,draw=blue!50,fill=blue!20,thick},
   dnn/.style={minimum size=6mm,rectangle,draw=black!50,fill=black!20,thick},
   simulator/.style={minimum size=20mm,rectangle,draw=black!50,fill=yellow!20,thick},
   arrow/.style={->,thick},
   simarrow/.style={->,thick,draw=red,dashed}]

\node [var] (st) {$s_t$};
\node [dnn] (pi) [above=of st]  {$\pi_\theta$};
\draw[arrow] (st) -- (pi);

\node [var] (at)  [above=of pi] {$a_t$};
\draw[arrow] (pi) -- (at);

\node [dnn] (DNNn) [right=of pi]  {$\DNN_N$};
\draw[arrow] (st) to [bend right=35] (DNNn.south);
\draw[arrow] (at) to [bend left=35]  (DNNn.north);

\node[dnn] (DNNr) [above=of at] {$\DNN_r$};
\draw[arrow] (at) -- (DNNr);
\node[var] (hrt) [above=of DNNr] {$\hat{r}_t$};
\draw[arrow] (DNNr) -- (hrt);
\draw[arrow] (st.west) to [bend left=35] (DNNr.west);

\node [var] (hstp)  [right=of DNNn] {$\hat{s}_{t+1}$};
\draw[arrow] (DNNn) -- (hstp);
\node [dnn] (plus)  [right=of hstp] {$+$};
\node [var] (nut)  [below=of plus,yshift=-5mm] {$\nu_{t+1}$};
\draw[arrow] (hstp.east) to (plus.west);
\draw[arrow] (nut.north) to (plus.south);

\node [var] (stp)  [right=of st,xshift=60mm] {$s_{t+1}$};
\draw[arrow] (plus.east) to ([xshift=-3mm] stp.north);

\node[simulator] (simtp) [below=of hstp,xshift=-10mm,yshift=-20mm] {Simulator$_{t+1}$};
\draw[simarrow] (hstp.south) to ([xshift=3mm,yshift=1mm] simtp.north);
\draw[simarrow] ([xshift=2mm] at.south) to ([yshift=1mm,xshift=1mm] simtp.north west);
\draw[simarrow] (simtp.east) to ([xshift=-2mm] nut.south);

\node[simulator] (simt) [left=of simtp,xshift=-35mm] {Simulator$_{t}$};
\draw[simarrow] (simt.east) to (simtp.west);
\node[simulator] (simtpp) [right=of simtp,xshift=35mm] {Simulator$_{t+2}$};
\draw[simarrow] (simtp.east) to (simtpp.west);

\begin{pgfonlayer}{background}
    \node [fill=black!30,fit=(st) (hrt) (simtp) (nut)] {};
  \end{pgfonlayer}

\end{tikzpicture}
\end{center}

Recall that we assume that $\hat{r}_t \approx r_t$ and $\hat{s}_{t+1} \approx s_{t+1}$. If these approximations are exact, then the dashed arrows can be eliminated from the figure above and the entire process of generating the rewards becomes a differentiable recurrent neural network. In such case, we can solve \eqref{eqn:policyOptimization} using the SGD framework, and at each iteration we calculate the gradient by  backpropagation in time over the recurrent neural network. 

In most situations, we expect $\hat{r}_t$ and $\hat{s}_{t+1}$ to slightly deviate from $r_t$ and $s_{t+1}$. The deviation of $\hat{r}_t$ from $r_t$ is less of an issue in practice, because it is often the case that there is nothing special about the exact reward $r_t$, and maximizing an approximation of it leads to similar performance. Therefore, for the sake of simplicity, we assume that maximizing the sum of $\hat{r}_t$ is sufficiently good. 

The more problematic part is the deviation of $\hat{s}_{t+1}$ from $s_{t+1}$. There are several possible sources for this deviation. 
\begin{enumerate}
\item \emph{Non-determinism}: in the traditional MDP model, $s_{t+1}$ is a random variable whose distribution is a function of $(s_t,a_t)$. But, the actual value of $s_{t+1}$ is not necessarily a deterministic function of $(s_t,a_t)$. 
\item \emph{Non-Markovianity}: it may be the case that the process is not Markovian in the state representation. It will always be Markovian in another representation, that is known to the simulator, but we do not know the Markovian representation or we do not want to model it. For example, in the ACC problem given in the next section, $s_{t+1}$ depends on the acceleration commands of the driver in front of us. While the simulator models this behavior in some complicated way, we do not want to model it and prefer to stick with a simple state representation that does not allow us to predict the acceleration of the other driver, but only allows us to react to the other driver's behavior. 
\item \emph{Failures of the learning process}: as we will show, we are going to learn $\DNN_N$ from examples, and we may suffer from the usual inaccuracies of learning algorithms (approximation error, estimation error, and optimization error). As this part is standard we ignore this issue and assume that we have managed to learn $\DNN_N$ sufficiently good. 
\end{enumerate}

In any case, despite the fact that $\hat{s}_{t+1}$ can deviate from $s_{t+1}$, we can still apply backpropagation in time in order to calculate an approximate gradient of the cumulative reward w.r.t. $\pi$. In particular, the forward part of the backpropagation is correct, due to the correction made by defining $s_{t+1}$ as a sum of the prediction $\hat{s}_{t+1}$ and the correction term $\nu_{t+1}$ (supplied by the simulator during the forward pass). In the backward part, we propagate the error through the solid arrows given in the figure, but we kill the messages that go through dashed arrows, because we refer to the simulator as a black box. As mentioned previously, we do not impose explicit probabilistic assumptions on $\nu_t$. In particular, we do not require Markovian relation. Instead, we rely on the recurrent network to propagate ``enough'' information between past and future through the solid arrows. Intuitively, $\DNN_N(s_t,a_t)$ describes the predictable part of the near future, while $\nu_t$ expresses the unpredictable aspects, mainly due to the behavior of other players in the environment. The learner should learn a policy that will be robust to the behavior of other players. Naturally, if $\|\nu_t\|$ is large, the connection between our past actions and future reward values will be too noisy for learning a meaningful policy. 

As noted in \cite{schmidhuber1991reinforcement}, explicitly expressing the dynamic of the system in a transparent way enables to incorporate prior knowledge more easily. For example, in \secref{sec:experiments} we demonstrate how prior knowledge greatly simplifies the problem of defining $\DNN_N$.

Finally, it is left to explain how we can learn $\DNN_r$ and $\DNN_N$ within the SL framework.
For this, we observe that by relying on the access to the simulator, we have the ability to generate tuples $(s,a,r,s')$ as training examples, where $s$ is the current state, $a$ is the action, $r$ is the reward, and $s'$ is the next state. We note that it is customary to use some elements of exploration in generating the training set. Since this is a standard technique, we omit the details. 
Equipped with such training examples, we can learn $\DNN_r$ in the SL framework by extracting examples of the form $((s,a),r)$ from each tuple $(s,a,r,s')$. The key point here is that even though the action $a$ is not necessarily optimal for $s$, it does not pose any problem for the task of learning the mapping from state-action into the correct reward. Furthermore, even though the reward is given in a ``bandit'' manner for the policy learning problem, it forms a ``full information'' feedback for the problem of learning a network $\DNN_r$, such that $\DNN_r(s_t,a_t) \approx r_t$. Likewise, we can learn $\DNN_N$ in  the SL framework by extracting examples of the form $((s,a),s')$ from each tuple $(s,a,r,s')$. Again, the fact that $a$ is not the optimal action for $s$ does not pose any problem for the task of learning the near future, $s'$, from the current state and action, $(s,a)$. 

It is also possible to simultaneously learn $\DNN_r, \DNN_N, $ and $\pi_\theta$, by defining an objective that combines the cumulative reward with supervised losses of the form $\|\hat{s}_{t+1}-s_{t+1}\|^2$ and $(\hat{r}_{t+1}-r_t)^2$. In the experimental section we report results for both separate and join training. 

\subsection{Robustness to Adversarial Environment}

Since our model does not impose probabilistic assumptions on $\nu_t$, we can consider environments in which $\nu_t$ is being chosen in an adversarial manner. Of course, we must make some restrictions on $\nu_t$, otherwise the adversary can make the planning problem impossible. A natural restriction is to require that $\|\nu_t\|$ is bounded by a constant.  Robustness against adversarial environment is quite useful in autonomous driving applications. We describe a real world aspect of adversarial environment in \secref{sec:experiments}.

Here, we show that choosing $\nu_t$ in an adversarial way might even speed up the learning process, as it can focus the learner toward the robust optimal policy.  We consider the following simple game. The state is $s_t \in \reals$, the action is $a_t \in \reals$, and the immediate loss function is $0.1 |a_t| + [|s_t| - 2]_+$, where $[x]_+ = \max\{x,0\}$ is the ReLU function. The next state is $s_{t+1} = s_t + a_t + \nu_t$, where $\nu_t \in [-0.5,0.5]$ is chosen by the environment in an adversarial manner.

It is possible to see that the optimal policy can be written as a two layer network with ReLU: $a_t = -[s_t - 1.5]_+ + [-s_t - 1.5]_+$. Observe that when $|s_t| \in (1.5,2]$, the optimal action has a larger immediate loss than the action $a=0$. Therefore, the learner must plan for the future and cannot rely solely on the immediate loss. 

Observe that the derivative of the loss w.r.t. $a_t$ is $0.1 
\,\sign(a_t)$ and the derivative w.r.t. $s_t$ is $1[|s_t| > 2] \, \sign(s_t)$. 
Suppose we are in a situation in which $s_t \in (1.5,2]$. The adversarial choice of $\nu_t$ would be to set $\nu_t = 0.5$, and therefore, we will have a non-zero loss on round $t+1$, whenever $a_t > 1.5 - s_t$. In all such cases, the derivative of the loss will back-propagate directly to $a_t$. We therefore see that the adversarial choice of $\nu_t$ helps the learner to get a non-zero back-propagation message in all cases for which the choice of $a_t$ is sub-optimal. 

\section{Example Applications} \label{sec:experiments}

The goal of this section is to demonstrate some aspects of our approach on two toy examples: adaptive cruise control (ACC) and merging into a roundabout. 

\subsection{The ACC Problem}

In the ACC problem, a host vehicle is trying to keep an adequate distance of 1.5 seconds to a target car, while driving as smooth as possible. We provide a simple model for this problem as follows.
The state space is $\reals^3$ and the action space is $\reals$. The first coordinate of the state is the speed of the target car, the second coordinate is the speed of the host car, and the last coordinate is the distance between host and target (namely, location of the host minus location of the target on the road curve). The action to be taken by the host is the acceleration, and is denoted by $a_t$. We denote by $\tau$ the difference in time between consecutive rounds (in the experiment we set $\tau$ to be 0.1 seconds). 

Denote $s_t = (v^{\textrm{target}}_t , v^{\textrm{host}}_t, x_t)$ and denote by $a^{\textrm{target}}_t$ the (unknown) acceleration of the target. The full dynamics of the system can be described by:
\begin{align*}
v^{\textrm{target}}_t &= [v^{\textrm{target}}_{t-1} + \tau\,a^{\textrm{target}}_{t-1} ]_+\\
v^{\textrm{host}}_t &= [v^{\textrm{host}}_{t-1} + \tau\,a_{t-1} ]_+ \\
x_t &= [x_{t-1} + \tau\,(v^{\textrm{target}}_{t-1} - v^{\textrm{host}}_{t-1} ) ]_+
\end{align*}
This can be described as a sum of two vectors:
\begin{align*}
s_t &= ( [s_{t-1}[0] + \tau a^{\textrm{target}}_{t-1}]_+ , [s_{t-1}[1] + \tau a_{t-1}]_+ , [s_{t-1}[2] + \tau(s_{t-1}[0]-s_{t-1}[1])]_+ ) \\
&= \underbrace{( s_{t-1}[0] , [s_{t-1}[1] + \tau a_{t-1}]_+ , [s_{t-1}[2] + \tau(s_{t-1}[0]-s_{t-1}[1])]_+ )}_{\DNN_N(s_{t-1},a_t)} + \underbrace{([s_{t-1}[0] + \tau a^{\textrm{target}}_{t-1}]_+ - s_{t-1}[0] , 0 , 0)}_{\nu_t} 
\end{align*}
The first vector is the predictable part and the second vector is the unpredictable part.

The reward on round $t$ is defined as follows:
\[
-r_t ~=~ 0.1 \, |a_t| ~+~ [ | x_t / x^*_t - 1 | - 0.3 ]_+ ~~~~\textrm{where}~~~~ x^*_t = \max\{1, 1.5\, v^{\textrm{host}}_t\}
\]
The first term above penalizes for any non-zero acceleration, thus encourages smooth driving. The second term depends on the ratio between the distance to the target car, $x_t$, and the desired distance, $x^*_t$, which is defined as the maximum between a distance of $1$ meter and brake distance of 1.5 seconds. Ideally, we would like this ratio to be exactly $1$, but as long as this ratio is in $[0.7,1.3]$ we do not penalize the policy, thus allowing the car some slack (which is important for maintaining a smooth drive). 

\subsection{Merging into a Roundabout}

In this experiment, the goal of the agent is to pass a roundabout. 
An episode starts when the agent is approaching the bottom entrance of the roundabout. The episode ends when the agent reaches the second exit of the roundabout, or after a fixed number of steps. A successful episode is measured first by keeping a safety distance from all other vehicles in the roundabout at all times. Second, the agent should finish the route as quickly as possible. And third, it should adhere a smooth acceleration policy. At the beginning of the episode, we randomly place $N_T$ target vehicles on the roundabout. 

To model a blend of adversarial and typical behavior, with probability $p$, a target vehicle is modeled by an ``aggressive'' driving policy, that accelerates when the host tries to merge in front of it. With probability $1-p$, the target vehicle is modeled by a ``defensive'' driving policy that deaccelerate and let the host merge in. In our experiments we set $p=0.5$. The agent has no information about the type of the other drivers. These are chosen at random at the beginning of the episode. 

We represent the state as the velocity and location of the host (the agent), and the locations, velocities, and accelerations of the target vehicles. Maintaining target accelerations is vital in order to differentiate between aggressive and defensive drivers based on the current state.  
All target vehicles move on a one-dimensional curve that outlines the roundabout path. The host vehicle moves on its own one-dimensional curve, which intersects the targets' curve at the merging point, and this point is the origin of both curves. To model reasonable driving, the absolute value of all vehicles' accelerations are upper bounded by a constant. Velocities are also passed through a ReLU because driving backward is not allowed. 
Note that by not allowing driving backward we make long-term planning a necessity (the agent cannot regret on its past action). 

\begin{figure}
\begin{center}
\includegraphics[width=0.3\textwidth]{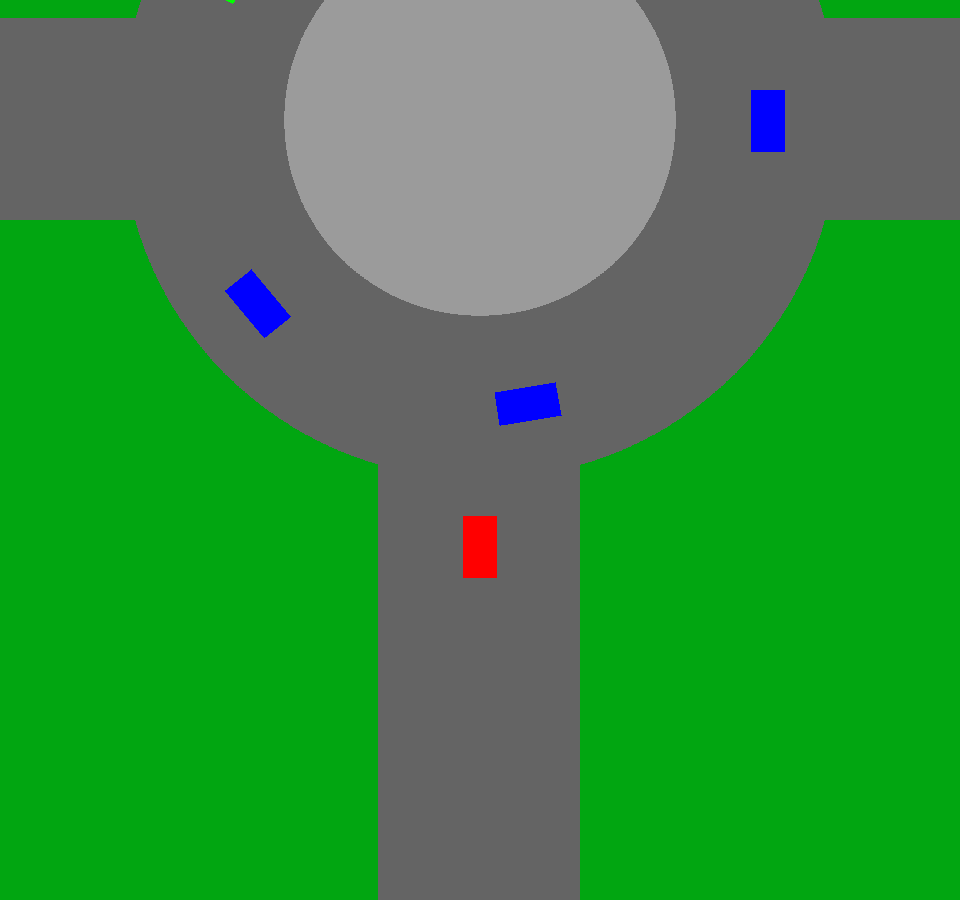} ~
\includegraphics[width=0.3\textwidth]{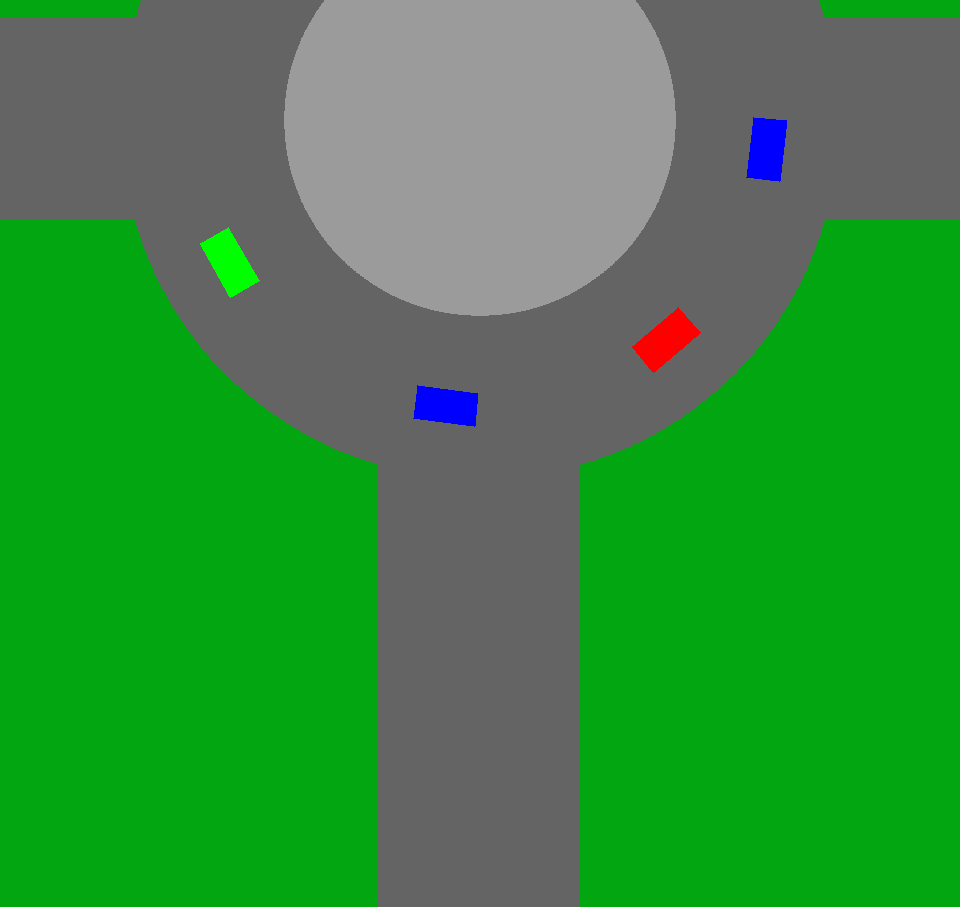}
\end{center}
\caption{Screenshots from the game. The agent is the car in red. Target vehicles are in blue (aggressive cars) and in green (defensive cars). The agent doesn't observe the type of the target cars. It should infer it from their position and acceleration. On the left, the agent correctly understands that the approaching car is aggressive and therefore stop and wait. On the right we see a successful merge.} \label{fig:demo}
\end{figure}

Recall that we decompose the next state, $s_{t+1}$, into a sum of a predictable part, $\DNN_N(s_t,a_t)$, and a non-predictable part, $\nu_{t+1}$. In our first experiment, we let $\DNN_N(s_t,a_t)$ be the dynamics of locations and velocities of all vehicles (which are well defined in a differentiable manner), while $\nu_{t+1}$ is the targets' acceleration. 
It is easy to verify that $\DNN_N(s_t,a_t)$ can be expressed as a combination of ReLU functions over an affine transformation, hence it is differentiable with respect to $s_t$ and $a_t$. The vector $\nu_{t+1}$ is defined by a simulator in a non-differentiable manner, and in particular implement aggressive behavior for some targets and defensive behavior for other targets. Two frames from the simulator are show in Figure~\ref{fig:demo}. As can be seen in the supplementary videos\footnote{\url{http://www.mobileye.com/mobileye-research/long-term-planning-by-short-term-prediction/}}, the agent learns to slowdown as it approaches the entrance of the roundabout. It also perfectly learned to give way to aggressive drivers, and to safely continue when merging in front of defensive ones. 

Our second experiment is more ambitious: we do not tell the network the function $\DNN_N(s_t,a_t)$. Instead, we express $\DNN_N$ as another learnable part of our recurrent network. Besides the rewards for the policy part, we add a loss term of the form $\|\DNN_N(s_t,a_t) - s_{t+1}\|^2$, where $s_{t+1}$ is the actual next state as obtained by the simulator. That is, we learn the prediction of the near future, $\DNN_N$, and the policy that plan for the long term, $\pi_\theta$, concurrently. While this learning task is more challenging, as can be seen in the supplementary videos, the learning process still succeeds.

\section{Discussion}
We have presented an approach for learning driving polices in the presence of other adversarial cars using recurrent neural networks. Our approach relies on partitioning of the near future into a predictable part and an un-predictable part. We demonstrated the effectiveness of the learning procedure for two simple tasks: adaptive cruise control and roundabout merging. The described technique can be adapted to learning driving policies in other scenarios, such as lane change decisions, highway exit and merge, negotiation of the right of way in junctions, yielding for pedestrians, as well as complicated planning in urban scenarios. 

\bibliographystyle{plainnat}
\bibliography{bib}

\begin{thebibliography}{29}
\providecommand{\natexlab}[1]{#1}
\providecommand{\url}[1]{\texttt{#1}}
\expandafter\ifx\csname urlstyle\endcsname\relax
  \providecommand{\doi}[1]{doi: #1}\else
  \providecommand{\doi}{doi: \begingroup \urlstyle{rm}\Url}\fi

\bibitem[Bakker(2001)]{bakker2001reinforcement}
Bram Bakker.
\newblock Reinforcement learning with long short-term memory.
\newblock In \emph{NIPS}, pages 1475--1482, 2001.

\bibitem[Bellman(1956)]{bellman1956dynamic}
Richard Bellman.
\newblock Dynamic programming and lagrange multipliers.
\newblock \emph{Proceedings of the National Academy of Sciences of the United
  States of America}, 42\penalty0 (10):\penalty0 767, 1956.

\bibitem[Bellman(1971)]{bellman1971introduction}
Richard Bellman.
\newblock \emph{Introduction to the mathematical theory of control processes},
  volume~2.
\newblock IMA, 1971.

\bibitem[Bertsekas(1995)]{bertsekas1995dynamic}
Dimitri~P Bertsekas.
\newblock \emph{Dynamic programming and optimal control}, volume~1.
\newblock Athena Scientific Belmont, MA, 1995.

\bibitem[Brafman and Tennenholtz(2003)]{brafman2003r}
Ronen~I Brafman and Moshe Tennenholtz.
\newblock R-max--a general polynomial time algorithm for near-optimal
  reinforcement learning.
\newblock \emph{The Journal of Machine Learning Research}, 3:\penalty0
  213--231, 2003.

\bibitem[Brown(1951)]{brown1951iterative}
George~W Brown.
\newblock Iterative solution of games by fictitious play.
\newblock \emph{Activity analysis of production and allocation}, 13\penalty0
  (1):\penalty0 374--376, 1951.

\bibitem[Cesa-Bianchi and Lugosi(2006)]{CesaBianchiLu06}
N.~Cesa-Bianchi and G.~Lugosi.
\newblock \emph{Prediction, learning, and games}.
\newblock Cambridge University Press, 2006.

\bibitem[HART and MAS-COLELL(2000)]{hart2000simple}
S.~HART and A.~MAS-COLELL.
\newblock A simple adaptive procedure leading to correlated equilibrium.
\newblock \emph{Econometrica}, 68\penalty0 (5), 2000.

\bibitem[Hu and Wellman(2003)]{hu2003nash}
Junling Hu and Michael~P Wellman.
\newblock Nash q-learning for general-sum stochastic games.
\newblock \emph{The Journal of Machine Learning Research}, 4:\penalty0
  1039--1069, 2003.

\bibitem[Kaelbling et~al.(1996)Kaelbling, Littman, and
  Moore]{kaelbling1996reinforcement}
Leslie~Pack Kaelbling, Michael~L Littman, and Andrew~W Moore.
\newblock Reinforcement learning: A survey.
\newblock \emph{Journal of artificial intelligence research}, pages 237--285,
  1996.

\bibitem[Kearns and Singh(2002)]{kearns2002near}
Michael Kearns and Satinder Singh.
\newblock Near-optimal reinforcement learning in polynomial time.
\newblock \emph{Machine Learning}, 49\penalty0 (2-3):\penalty0 209--232, 2002.

\bibitem[Kober et~al.(2013)Kober, Bagnell, and Peters]{kober2013reinforcement}
Jens Kober, J~Andrew Bagnell, and Jan Peters.
\newblock Reinforcement learning in robotics: A survey.
\newblock \emph{The International Journal of Robotics Research}, page
  0278364913495721, 2013.

\bibitem[Laud(2004)]{laud2004theory}
Adam~Daniel Laud.
\newblock \emph{Theory and application of reward shaping in reinforcement
  learning}.
\newblock PhD thesis, University of Illinois at Urbana-Champaign, 2004.

\bibitem[Littman(1994)]{littman1994markov}
Michael~L Littman.
\newblock Markov games as a framework for multi-agent reinforcement learning.
\newblock In \emph{Proceedings of the eleventh international conference on
  machine learning}, volume 157, pages 157--163, 1994.

\bibitem[Mnih et~al.(2014)Mnih, Heess, Graves, et~al.]{mnih2014recurrent}
Volodymyr Mnih, Nicolas Heess, Alex Graves, et~al.
\newblock Recurrent models of visual attention.
\newblock In \emph{Advances in Neural Information Processing Systems}, pages
  2204--2212, 2014.

\bibitem[Mnih et~al.(2015)Mnih, Kavukcuoglu, Silver, Rusu, Veness, Bellemare,
  Graves, Riedmiller, Fidjeland, Ostrovski, et~al.]{mnih2015human}
Volodymyr Mnih, Koray Kavukcuoglu, David Silver, Andrei~A Rusu, Joel Veness,
  Marc~G Bellemare, Alex Graves, Martin Riedmiller, Andreas~K Fidjeland, Georg
  Ostrovski, et~al.
\newblock Human-level control through deep reinforcement learning.
\newblock \emph{Nature}, 518\penalty0 (7540):\penalty0 529--533, 2015.

\bibitem[Ng et~al.(1999)Ng, Harada, and Russell]{ng1999policy}
Andrew~Y Ng, Daishi Harada, and Stuart Russell.
\newblock Policy invariance under reward transformations: Theory and
  application to reward shaping.
\newblock In \emph{ICML}, volume~99, pages 278--287, 1999.

\bibitem[Sch{\"a}fer(2008)]{schafer2008reinforcement}
Anton~Maximilian Sch{\"a}fer.
\newblock \emph{Reinforcement Learning with Recurrent Neural Network}.
\newblock PhD thesis, Universitat Osnabruck, 2008.

\bibitem[Schmidhuber(1991)]{schmidhuber1991reinforcement}
J{\"u}rgen Schmidhuber.
\newblock Reinforcement learning in markovian and non-markovian environments.
\newblock In \emph{NIPS}, 1991.

\bibitem[Shoham et~al.(2007)Shoham, Powers, and Grenager]{shoham2007if}
Yoav Shoham, Rob Powers, and Trond Grenager.
\newblock If multi-agent learning is the answer, what is the question?
\newblock \emph{Artificial Intelligence}, 171\penalty0 (7):\penalty0 365--377,
  2007.

\bibitem[Silver et~al.(2014)Silver, Lever, Heess, Degris, Wierstra, and
  Riedmiller]{silver2014deterministic}
David Silver, Guy Lever, Nicolas Heess, Thomas Degris, Daan Wierstra, and
  Martin Riedmiller.
\newblock Deterministic policy gradient algorithms.
\newblock In \emph{ICML}, 2014.

\bibitem[Sutton and Barto(1998)]{sutton1998reinforcement}
Richard~S Sutton and Andrew~G Barto.
\newblock \emph{Reinforcement learning: An introduction}, volume~1.
\newblock MIT press Cambridge, 1998.

\bibitem[Szepesv{\'a}ri(2010)]{szepesvari2010algorithms}
Csaba Szepesv{\'a}ri.
\newblock Algorithms for reinforcement learning.
\newblock \emph{Synthesis Lectures on Artificial Intelligence and Machine
  Learning}, 4\penalty0 (1):\penalty0 1--103, 2010.
\newblock URL \url{http://www.ualberta.ca/~szepesva/RLBook.html}.

\bibitem[Thrun(1995)]{Thrun95a}
S.~Thrun.
\newblock Learning to play the game of chess.
\newblock In G.~Tesauro, D.~Touretzky, and T.~Leen, editors, \emph{Advances in
  Neural Information Processing Systems (NIPS) 7}, Cambridge, MA, 1995. MIT
  Press.

\bibitem[Uther and Veloso(1997)]{uther1997adversarial}
William Uther and Manuela Veloso.
\newblock Adversarial reinforcement learning.
\newblock Technical report, Technical report, Carnegie Mellon University, 1997.
  Unpublished, 1997.

\bibitem[Watkins and Dayan(1992)]{watkins1992q}
Christopher~JCH Watkins and Peter Dayan.
\newblock Q-learning.
\newblock \emph{Machine learning}, 8\penalty0 (3-4):\penalty0 279--292, 1992.

\bibitem[White~III(1991)]{white1991survey}
Chelsea~C White~III.
\newblock A survey of solution techniques for the partially observed markov
  decision process.
\newblock \emph{Annals of Operations Research}, 32\penalty0 (1):\penalty0
  215--230, 1991.

\bibitem[Williams(1992)]{williams1992simple}
Ronald~J Williams.
\newblock Simple statistical gradient-following algorithms for connectionist
  reinforcement learning.
\newblock \emph{Machine learning}, 8\penalty0 (3-4):\penalty0 229--256, 1992.

\bibitem[Xu et~al.(2015)Xu, Ba, Kiros, Courville, Salakhutdinov, Zemel, and
  Bengio]{xu2015show}
Kelvin Xu, Jimmy Ba, Ryan Kiros, Aaron Courville, Ruslan Salakhutdinov, Richard
  Zemel, and Yoshua Bengio.
\newblock Show, attend and tell: Neural image caption generation with visual
  attention.
\newblock \emph{arXiv preprint arXiv:1502.03044}, 2015.

\end{thebibliography}

\end{document}